\documentclass[10pt,twocolumn,letterpaper]{article}

\usepackage[pagenumbers]{cvpr} 

\usepackage{geometry}
\usepackage{tabularx}
\usepackage{booktabs}
\usepackage{array}
\usepackage{multirow}
\usepackage[T1]{fontenc}
\PassOptionsToPackage{defaults=hu-min,classmod=unchanged}{magyar.ldf}
\usepackage[magyar,english]{babel}
\definecolor{cvprblue}{rgb}{0.21,0.49,0.74}
\usepackage[pagebackref,breaklinks,colorlinks,allcolors=cvprblue]{hyperref}

\def\paperID{FNfl8Ceh01} 
\def\confName{CVPR}
\def\confYear{2025}

\title{Synthetic Document Question Answering in Hungarian}

\author{
Jonathan Li\textsuperscript{1} \and%
Zoltan Csaki\textsuperscript{1} \and%
Nidhi Hiremath\textsuperscript{1} \and%
Etash Guha\textsuperscript{1,2} \and%
Fenglu Hong\textsuperscript{1} \and%
Edward Ma\textsuperscript{1} \and%
Urmish Thakker\textsuperscript{1}%
\\%
\\%
\textsuperscript{1}SambaNova Systems, Inc.\quad%
\textsuperscript{2}University of Washington%
}%
\begin{document}
\maketitle%
\begin{abstract}
Modern VLMs have achieved near-saturation accuracy in English document visual question-answering (VQA). However, this task remains challenging in lower resource languages due to a dearth of suitable training and evaluation data. In this paper we present scalable methods for curating such datasets by focusing on Hungarian, approximately the 17th highest resource language on the internet \cite{nguyen2023culturax}. Specifically, we present HuDocVQA and HuDocVQA-manual,  document VQA datasets that modern VLMs significantly underperform on compared to English DocVQA \cite{mathew2021docvqadatasetvqadocument}. HuDocVQA-manual is a small manually curated dataset based on Hungarian documents from Common Crawl \cite{commoncrawl}, while HuDocVQA is a larger synthetically generated VQA data set from the same source. We apply multiple rounds of quality filtering and deduplication to HuDocVQA in order to match human-level quality in this dataset. We also present HuCCPDF, a dataset of 117k pages from Hungarian Common Crawl PDFs along with their transcriptions, which can be used for training a model for Hungarian OCR. To validate the quality of our datasets, we show how finetuning on a mixture of these datasets can improve accuracy on HuDocVQA for Llama 3.2 11B Instruct by +7.2\%. We release our \href{https://huggingface.co/collections/jlli/hungarian-document-datasets-682f24bf847b3c7376b493bc}{datasets} and \href{https://github.com/snova-jonathanl/HuDocVQA}{code} to foster further research in multilingual DocVQA.

\end{abstract}    
\section{Introduction}
\label{sec:intro}

\begin{figure}
    \centering
    \includegraphics[width=\linewidth]{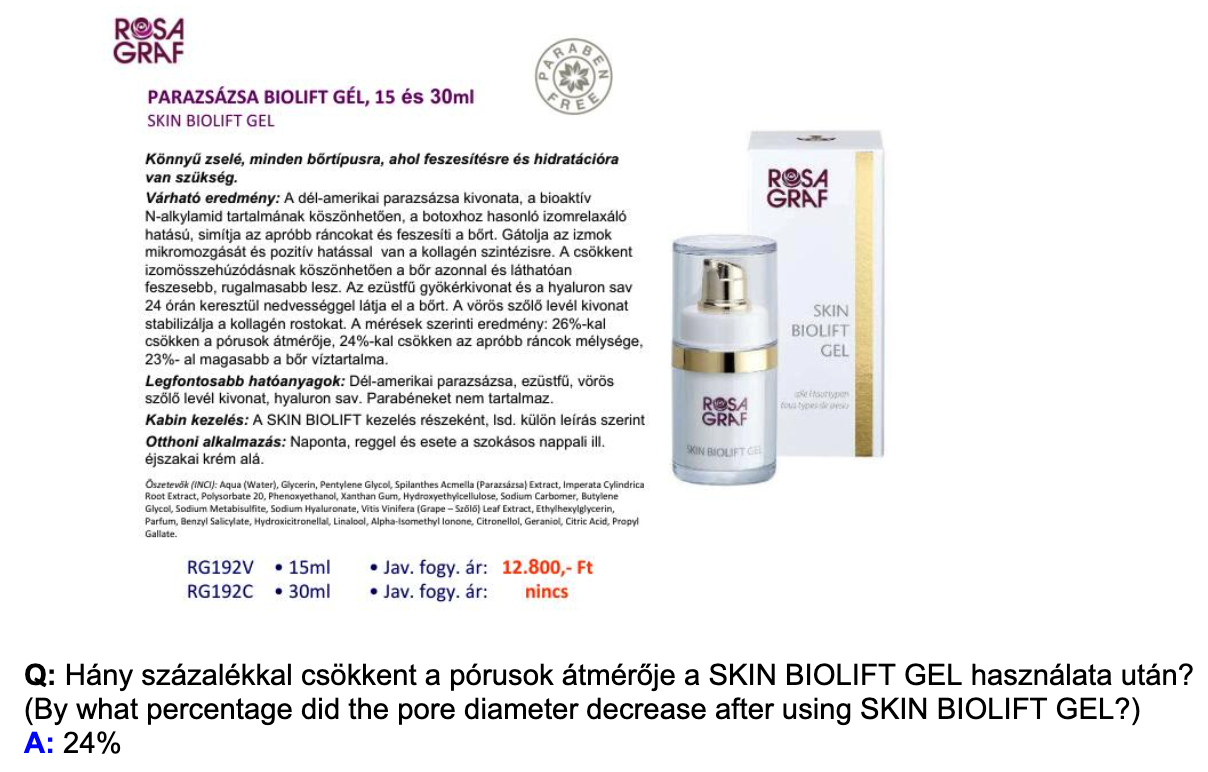}
    \caption{An example question-answer pair from HuDocVQA-manual}
    \label{fig:docvqa_example}
\end{figure}

As large language models (LLMs) quickly reach saturation and human-level performance on a host of traditional NLP tasks, vision-language models (VLMs) are quickly reaching saturation on traditional multimodal benchmarks as well. One such benchmark is DocVQA \cite{mathew2021docvqadatasetvqadocument}, where human-level performance is estimated at 94.36\% exact-match accuracy and 0.981 ANLS \cite{goyal2017makingvvqamatter}. Recent open-weights VLMs such as Llama 3.2 90B Instruct \cite{grattafiori2024llama3herdmodels} and Qwen 2.5 72B Instruct \cite{bai2025qwen2d5vl}, as well as closed-weights VLMs in GPT-4o and Claude 3.7 Sonnet have achieved well over 0.9 ANLS. 

However, while LLM training and evaluation has diversified and shown proficiency in a host of non-English and even low-resource languages~\cite{grattafiori2024llama3herdmodels,kolavi2024,shenzhi_wang_2024,qwen2025qwen25technicalreport,yue2024pangeafullyopenmultilingual}, there are comparatively fewer studies on the multilingual capabilities of VLMs. In particular, multilingual DocVQA suffers from a surprising lack of training and test data, with a major pain point being the scalability of dataset creation efforts. Existing multilingual VQA works fail to address these issues: EXAMS-V \cite{das2024examsv} and JDocQA \cite{onami-etal-2024-jdocqa-japanese} involved painstaking manual curation and quality checking by native speakers, while MaxM \cite{changpinyo2022maxm} applies machine translation to captioning data, which may suffer from ``translationese'' \cite{riley-etal-2020-translationese}. A more detailed discussion on related work can be found in Appendix \ref{sec:related}.

In this paper, we take advantage of the multilingual capabilities of LLMs to scalably construct multilingual DocVQA training and evaluation data. In Section \ref{sec:dataset}, we describe the dataset curation process, including sourcing, synthetic data generation, and quality filtering for 3 Hungarian multimodal datasets: HuDocVQA-manual, HuDocVQA, and HuCCPDF. In Section \ref{sec:evals}, we show how modern VLMs significantly underperform on HuDocVQA and HuDocVQA-manual compared to English DocVQA, and how finetuning on HuDocVQA and HuCCPDF can improve accuracy by up to +7.2\% for existing strong VLMs like Llama 3.2 11B Instruct. We plan to publicly release our datasets and code to further foster scalable construction of multilingual document datasets.
\section{Dataset Curation}
\label{sec:dataset}

\subsection{HuDocVQA-manual}
To create HuDocVQA-manual, the authors annotated 54 pages of Hungarian PDFs from Common Crawl to serve as a human-verified benchmark for document VQA in Hungarian. For each page, we used Google Translate and DeepL to read the document and create a semantically meaningful question-answer pair in English. Then, we translated the English pair back into Hungarian with the same tools. All examples were then verified by a native Hungarian speaker.

\subsection{HuDocVQA}
\begin{figure}[htbp]
  \centering
  \includegraphics[trim={0.5cm 2cm 5cm 2cm},clip,width=\linewidth]{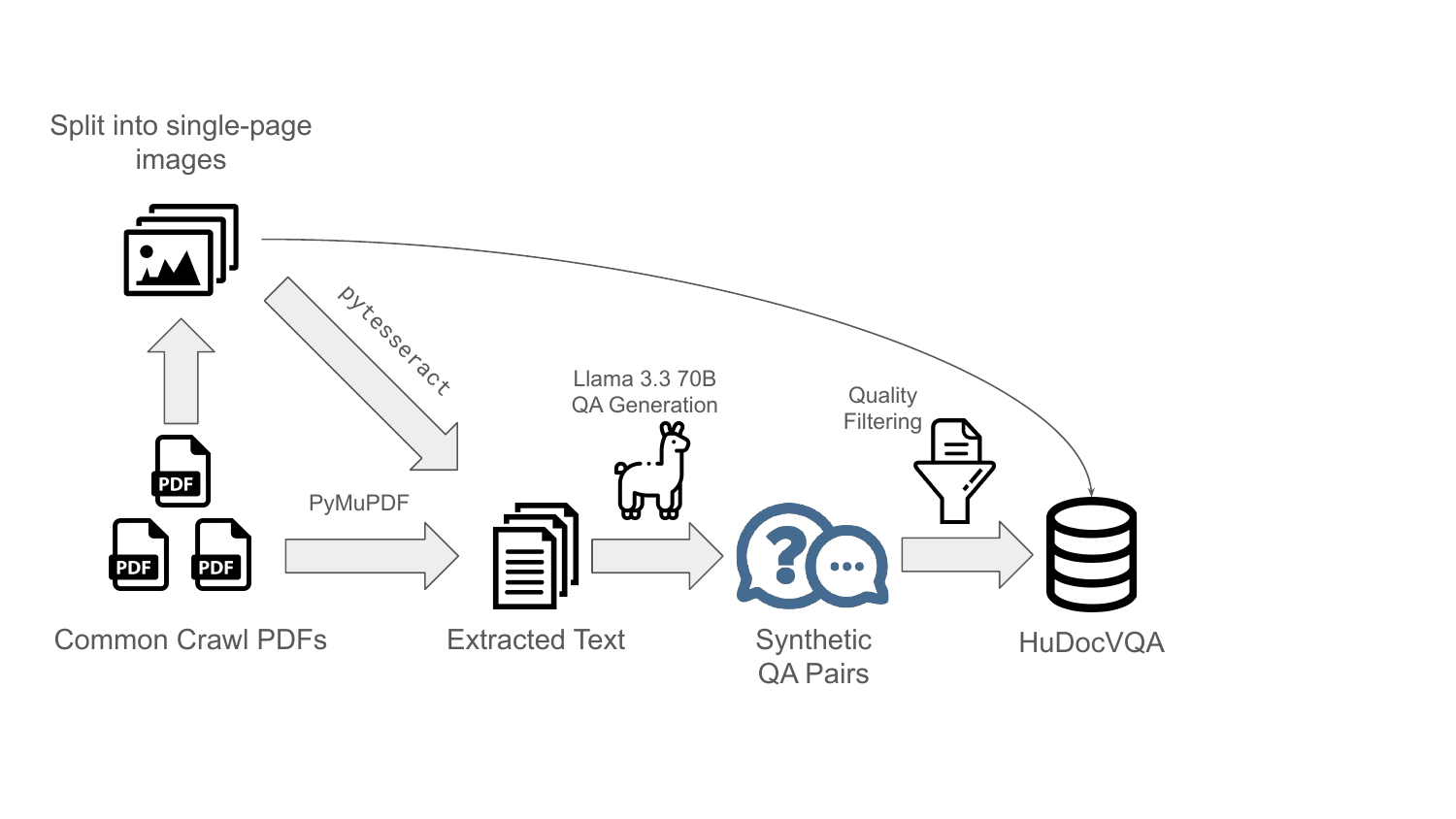}
  \caption{A diagram of our synthetic data pipeline}
  \label{fig:syndatapipeline}
\end{figure}

In this section, we describe each step of our synthetic question-answer generation pipeline for HuDocVQA. See \cref{fig:syndatapipeline} for a diagram of the full process.

\subsubsection{Synthetic VQA Generation}
\label{subsubsec:synqa_generation}

To generate synthetic VQA pairs, we start with a large collection of PDFs in the target language, as described in \cref{subsec:HuCCPDF}, totaling 24,341 pages of Hungarian text. We treat each page as a single image and apply two methods of text extraction: OCR using \texttt{pytesseract}~\citep{pytesseract} and PDF parsing with PyMuPDF~\citep{pymupdf2025}. This helps mitigate biases introduced by each text extraction method. 

Once text is extracted from a page, we construct a one-shot prompt using one of three randomly sampled human-written Hungarian QA pairs. We then use LLaMA 3.3 70B \citep{grattafiori2024llama3herdmodels} to generate new QA pairs based on the one-shot prompt and the extracted text. For each image, we generate 4 QA pairs (2 from the PyTesseract text and 2 from the PyMuPDF text) and discard any model outputs that do not follow the question and answer format of the few-shot examples. For generation and prompting details, see \cref{sec:synqa_gen_details}. In total, we generate 93,933 QA pairs, averaging 3.9 questions per image. In addition, we create training, test, and validation splits based on each image. See \cref{tab:dataset_statistics} for full dataset statistics.

\begin{table}[htbp]
    \centering
    \begin{tabular}{|p{2cm}|c|c|c|}
        \hline
        \textbf{Dataset} & \textbf{Split} & \textbf{\# Images} & \textbf{\# QAs} \\
        \hline
        HuDocVQA-manual& Test & 54 & 54 \\
        \hline
        \multirow{3}{*}{HuDocVQA} & Train & 21,800 & 84,192 \\
        \cline{2-4}
                                  & Validation & 1,283 & 4,930 \\
        \cline{2-4}
                                  & Test & 1,258 & 4,811 \\
        \hline
        \multirow{3}{*}{+ filtering} & Train & 20,059 & 55,325 \\
        \cline{2-4}
                                  & Validation & 1,183 & 3,382 \\
        \cline{2-4}
                                  & Test & 1,162 & 3,315 \\
        \hline
    \end{tabular}
    \caption{HuDocVQA Statistics}
    \label{tab:dataset_statistics}
\end{table}

\subsubsection{Filtering VQA Data}
\label{subsubsec:quality_filters}
In our initial observations of the generated synthetic VQA data, we identified several recurring issues. For each issue, we developed quality checks to filter out examples containing these issues. \cref{tab:filters} summarizes these common issues along with the heuristic filters applied to mitigate them, and  \cref{tab:suppl_filter_stats} shows the number of examples removed out by each heuristic filter. In total, we filter out 31,911 QA pairs, leaving 62,022 remaining to comprise HuDocVQA.

\begin{table}[ht]
\centering
\renewcommand{\arraystretch}{1.4}
\begin{tabularx}{\linewidth}{|>{\raggedright\arraybackslash}X|>{\raggedright\arraybackslash}X|}
\hline
\textbf{Issue} & \textbf{Filter} \\
\hline
The text in the image is too short (e.g., only a title). & Text length filter: text must exceed 60 characters \\
\hline
The question is not related to the text in the image. & N-gram overlap between question and text (n=4), with a 12\% overlap threshold \\
\hline
The question is generated in the wrong language. & Language filter using Python's \texttt{langdetect} \citep{danilak_langdetect} \\
\hline
Duplicate questions are generated for the same image. & Use LLM to detect whether any two questions for the same image are paraphrases \\
\hline
\end{tabularx}
\caption{Issues identified in our synthetic data and the corresponding filters we applied.}
\label{tab:filters}
\end{table}

\subsection{HuCCPDF}\label{subsec:HuCCPDF}

In order to address the shortage of high-quality datasets in Hungarian for both OCR and visual question answering, we filtered PDFs from Common Crawl\cite{commoncrawl} to create HuDocVQA-manual, HuDocVQA, and HuCCPDF. The same procedure was applied to collect PDFs for all 3 datasets, with image-level deduplication applied between all 3.

Specifically for HuCCPDF, we processed 3750/60,000 WARC \cite{warcformat} files from \href{https://commoncrawl.org/blog/april-2021-crawl-archive-now-available}{CC-MAIN-2021-17}. We collect links to PDFs from each WARC file, download them, and extract their content in plaintext and Markdown with PyMuPDF~\cite{pymupdf2025,pymupdf4llm}. We filter out non-Hungarian PDFs using fastText \citep{fasttext}, as well as pages with less than 100 characters of text and pages with significant differences in plaintext and Markdown. See \cref{sec:pdf_collection} for more details.

HuCCPDF consists of approximately 38,000 Hungarian PDFs, resulting in 113,091 pages after filtering. This constitutes 0.2\% of the total PDFs we collected, which is consistent with the distribution of Hungarian text in previous multilingual datasets~\cite{nguyen2023culturax}.

\section{Evaluations}
\label{sec:evals}

\subsection{Benchmarking State-of-the-Art VLMs}

\begin{table*}
  \centering
  \begin{tabularx}{\textwidth}{|X|X|X|X|}
    \hline
    \textbf{Model} & \textbf{DocVQA (val)} & \textbf{HuDocVQA-manual} & \textbf{HuDocVQA (test)} \\
    \hline
    GPT 4o & 0.815 & \textbf{0.667} & \textbf{0.694}\\
    Claude 3.7 Sonnet & 0.938 & 0.630 & 0.655\\
    Qwen 2.5 VL 72B & \textbf{0.961} & 0.611 & 0.613\\
    Llama 3.2 90B Instruct & 0.954 & 0.481 & 0.498 \\
    \hline
  \end{tabularx}
  \caption{Comparison of leading open-source and closed-source VLMs on DocVQA, HuDocVQA-manual, and HuDocVQA. Accuracy is measured with LLM-as-a-Judge.}
  \label{tab:sota_vlms}
\end{table*}

In \cref{tab:sota_vlms} we evaluate closed- and open-weights VLMs on HuDocVQA-manual and HuDocVQA, and compare their scores to their performance on DocVQA. Notably, accuracy is measured with LLM-as-a-Judge as opposed to ANLS \cite{goyal2017makingvvqamatter}, the typical DocVQA performance metric \cite{mathew2021docvqadatasetvqadocument}. See Appendix \cref{sec:suppl_anls} for more details justifying our approach.

While all models achieve near- or above human performance on the English task, all models underperform by around 30\% in the Hungarian equivalent. Llama 3.2 90B Instruct has a particularly noticeable drop in performance, at almost half of its English accuracy.

We also note that all models achieve similar scores on HuDocVQA-manual and HuDocVQA: the scores have a Pearson coefficient of 0.986 and a p-value of 0.01, suggesting that our synthetic data pipeline achieves comparable quality as human annotation. Given these results, we track accuracy for the HuDocVQA test set only in subsequent experiments.

\subsection{Finetuning Experiments}
\label{subsec:ft_experiments}

\begin{table}[ht]
  \centering
  \begin{tabularx}{\linewidth}{|l|X|}
    \hline
    \textbf{Model} & \textbf{HuDocVQA (test)} \\
    \hline
    Llama 3.2 11B Instruct & 0.332\\
    + HuDocVQA & 0.285 \\
    + SFT mixture & 0.303 \\
    + SFT + 21k OCR & 0.374 \\
    + SFT + 105k OCR & \textbf{0.404} \\
    \hline
  \end{tabularx}
  \caption{Llama 3.2 evaluations on HuDocVQA datasets. All accuracy numbers are computed with LLM-as-a-Judge.}
  \label{tab:llama_finetuning}
\end{table}

\cref{tab:llama_finetuning} shows the evaluation results of our finetuning experiments on Llama 3.2 11B Instruct. We can see that finetuning on HuDocVQA alone leads to a drop in accuracy of -4.7\%.  We hypothesize this drop in accuracy was due to insufficient data scale and variety; to remedy this, we finetune on a mixture of the Cauldron
\cite{laurençon2024mattersbuildingvisionlanguagemodels}, Docmatix \cite{laurençon2024buildingbetterunderstandingvisionlanguage}, LAION-COCO-NLLB \cite{visheratin2023nllbcliptrainperformant} and HuDocVQA, indicated by ``SFT mixture'' (see \cref{sec:suppl_sft_mixture} for more details). The resulting model achieves an accuracy of 0.303, improving on the single-task run but still underperforming compared to the baseline. Finally, we add 105k examples from HuCCPDF to the mixture, achieving a final accuracy of 0.404 on HuDocVQA. These examples are formatted as an OCR task: given an image of a single page, predict the ground-truth text.

For all experiments, we apply simple model averaging \cite{wortsman2022modelsoupsaveragingweights} to obtain the final checkpoint, which we observe provides a small boost to all scores (see \cref{sec:merge_ablation} for more details).


\subsection{Finetuning Ablations}

In \cref{tab:llama_finetuning}, we ablate the amount of OCR data from HuCCPDF we add to the SFT mixture. We observe that adding 21k datapoints from HuCCPDF improves the final accuracy to 4.2\% above the baseline, and adding 105k datapoints increases accuracy further to 7.2\% above the baseline. We also run ablations on the application of quality filters in \cref{subsubsec:quality_filters} and model merging to justify our approach. See \cref{sec:suppl_ablations} for more details.

\section{Conclusion}
\label{sec:conclusion}
In summary, we present 3 multimodal document datasets in Hungarian: HuDocVQA, HuDocVQA-manual, and HuCCPDF. We describe our dataset curation and synthetic data generation steps in detail, utilizing the multilingual capabilities of LLMs to scalably construct QA pairs for Hungarian PDFs and applying careful quality filtering of their outputs. We demonstrate how state-of-the-art VLMs underperform on the DocVQA task in Hungarian compared to English, and our training experiments indicate further training on Hungarian document data can boost performance. Our experiments and methodologies can be applied to any language, and we hope this work can encourage high-quality multilingual SFT datasets for future VLMs.

{
    \small
    \bibliographystyle{ieeenat_fullname}
    \bibliography{main}
}

\clearpage
\setcounter{page}{1}
\maketitlesupplementary

\section{Related Works}
\label{sec:related}

\subsection{DocVQA}
DocVQA \cite{mathew2021docvqadatasetvqadocument} is the seminal dataset and benchmark on the Document Visual Question Answering task. Since the release of LLaVA-NeXT \cite{liu2024llavanext}, most multimodal model releases have benchmarked their OCR abilities on this dataset. At the time of writing, multiple open- and closed-source models \cite{chen2024expandinginternvl2d5}\cite{bai2025qwen2d5vl} \cite{openai2024gpt4ocard} have claimed to perform at or above human-level performance of 94.4\% accuracy, which we have verified in our own benchmarking.

With respect to multilingual VQA, the MaXM dataset \cite{changpinyo2022maxm} consists of visual QA pairs in 7 languages, synthetically generated from the XM-3600 dataset \cite{thapliyal2022crossmodal} using mT5-XXL \cite{xue2021mt5massivelymultilingualpretrained}. However, this dataset is limited by both the simplicity of its QA pairs (target answers are often simple descriptions of the image) and the lack of native document data.

JDocQA \cite{onami-etal-2024-jdocqa-japanese} addresses both issues, in presenting a large-scale, human-annotated document QA dataset in Japanese. The dataset includes open-ended questions that cannot be answered by repeating a piece of text in the image verbatim, and as such presents a more difficult task than DocVQA. HuDocVQA, in contrast, is synthetically generated and requires much less manual labor to collect while still maintaining human-level annotation quality.

\subsection{Visual Synthetic Data}

MaXM \cite{changpinyo2022maxm} presented early promising signs of using text-only LLMs to generate synthetic multilingual text data for images. Modern VLMs, such as Llama 3.2 \cite{grattafiori2024llama3herdmodels} and Qwen 2.5 VL \cite{bai2025qwen2d5vl}, also note the use of text-only LLMs to generate synthetic QA or conversational post-training data for VLMs. Qwen 2.5 VL, in particular, makes note of rule- and model-based filtering techniques for their synthetic visual data.

\subsection{Multilingual VLMs}

A bevy of frontier VLMs, including Llama 3.2 \cite{grattafiori2024llama3herdmodels}, Mistral Small 3.2 \cite{mistralsmall3d1}, Gemma 3 and Qwen 2.5 VL \cite{bai2025qwen2d5vl} claim to be multilingual, while only Qwen specifically calls out inclusion of multilingual OCR data during visual pre-training. PALO \cite{maaz2024palopolyglotlargemultimodal} presented a recipe to construct a natively multilingual VLM from CLIP ViT-L \cite{radford2021learningtransferablevisualmodels} and Vicuna models \cite{zheng2023judgingllmasajudgemtbenchchatbot}, using GPT-3.5-Turbo and language-specific scripts to translate Llava pretraining and finetuning datasets into multiple languages. Notably, PALO did not target OCR capabilities in their multilingual translation efforts.

\section{Filtering Statistics}
\label{sec:filter_stats}

\begin{table}[ht]
\centering
\renewcommand{\arraystretch}{1.4}
\begin{tabularx}{\linewidth}{|>{\raggedright\arraybackslash}X|>{\raggedright\arraybackslash}X|}
\hline
\textbf{Filter} & \textbf{\# QAs filtered} \\
\hline
text length $>$ 60 & 856 (0.9\%)\\
\hline
$n$-gram overlap & 7,931 (8.5\%)\\
\hline
\texttt{langdetect} & 2,622 (3.1\%) \\
\hline
deduplication & 20,393 (24.7\%) \\
\hline
\end{tabularx}
\caption{Number of synthetic QAs filtered out for each of the filters applied}
\label{tab:suppl_filter_stats}
\end{table}

\cref{tab:suppl_filter_stats} shows the numbers of question-answer pairs filtered from HuDocVQA for each of the heuristic filters described in \cref{tab:filters}. Note that percentages are taken with respect to the input dataset to each filter, so the 8.5\% removed from $n$-gram overlap is calculated after removing 856 examples due to text length.

\section{ANLS vs LLM-as-a-Judge}
\label{sec:suppl_anls}

All accuracy measurements on all datasets are measured with LLM-as-a-Judge \cite{zheng2023judgingllmasajudgemtbenchchatbot}, using Llama 3.1 405B Instruct as the judge. While ANLS serves as a fine metric for English DocVQA, in early experiments with HuDocVQA-manual that it often went against human judgment on model response correctness. In particular, training experiments would always show ANLS increasing, despite accuracy sometimes degrading (as in \cref{tab:filtering_ablation}). As such, we decided to use LLM-as-a-Judge as a proxy for human correct judgements.

\begin{figure*}
    \centering
    \includegraphics[width=\textwidth]{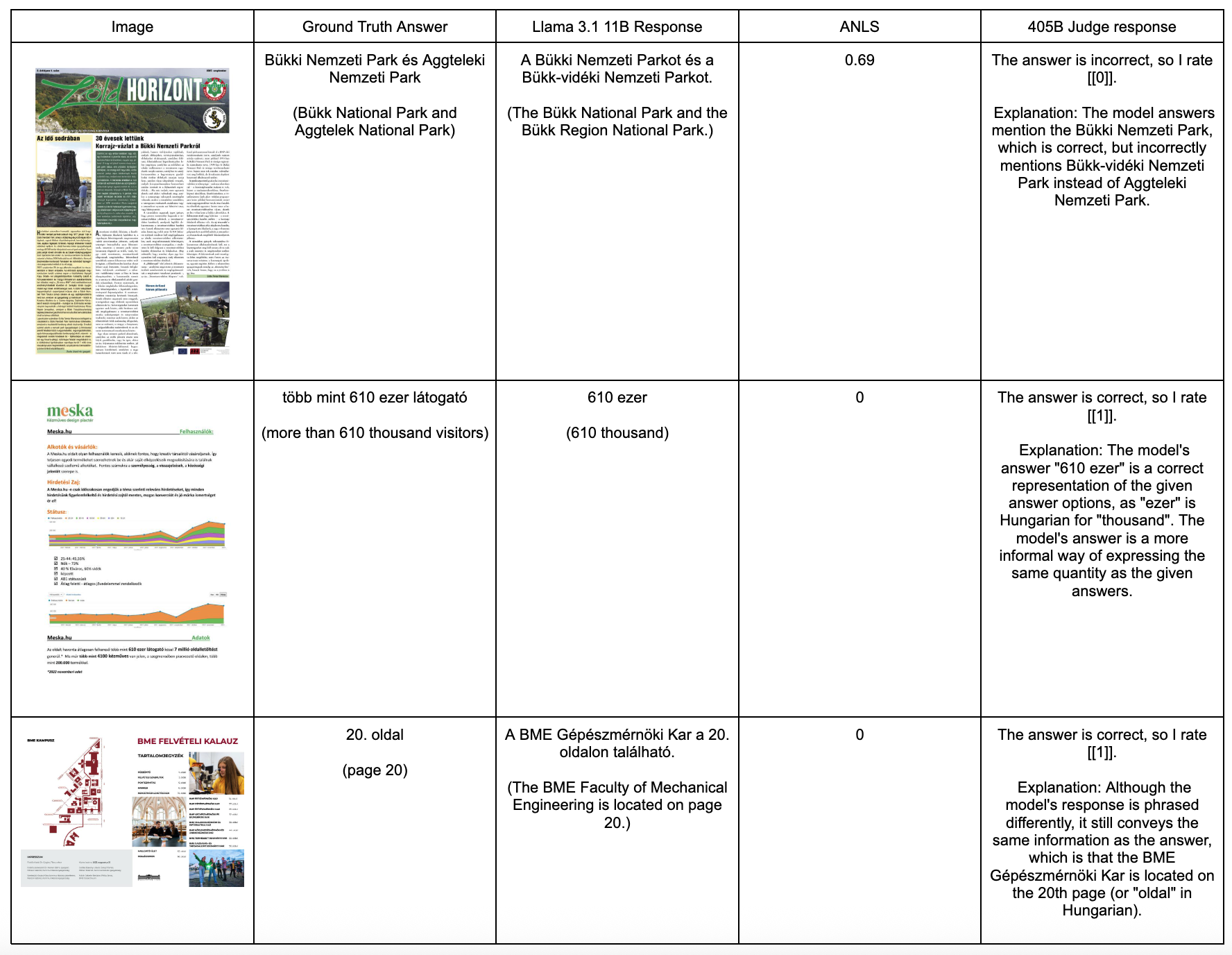}
    \caption{Examples from HuDocVQA-manual where ANLS fails to correctly measure model response accuracy in HuDocVQA-manual}
    \label{fig:anls_examples}
\end{figure*}

\begin{figure*}
    \centering
    \includegraphics[width=\textwidth]{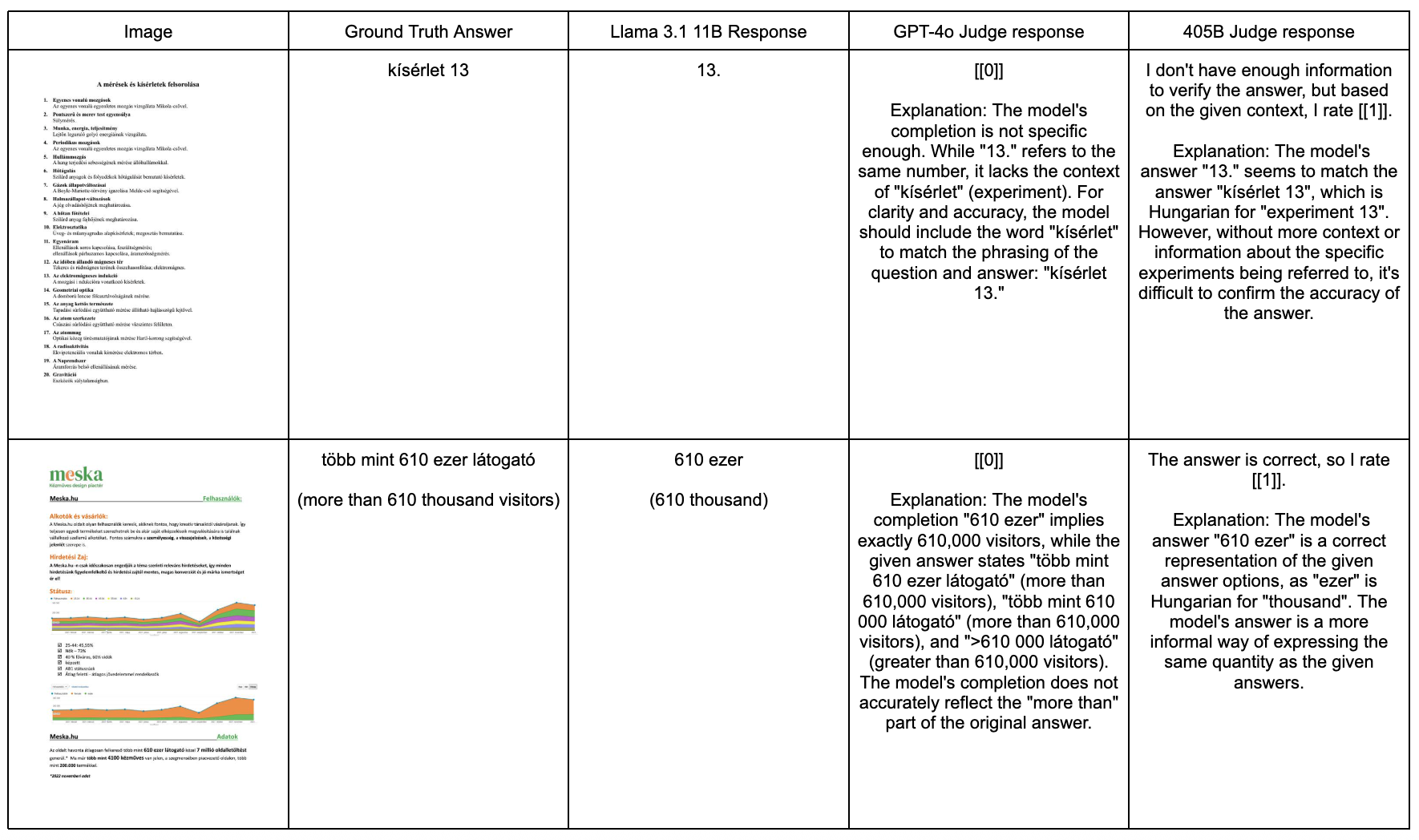}
    \caption{Examples from HuDocVQA-manual where GPT-4o would rate models more harshly than Llama 3.1 405B Instruct}
    \label{fig:gpt4_v_405b_examples}
\end{figure*}

\cref{fig:anls_examples} show such examples from HuDocVQA-manual, where ANLS might deviate from human judgements of correctness of model responses, while Llama 3.1 405B Instruct's judgment aligns better. We had also experimented with using GPT-4o as a judge, but found that GPT-4o would rate model responses more harshly than a human would, while Llama 3.1 405B Instruct aligned with our judgments better. See \cref{fig:gpt4_v_405b_examples} for examples from HuDocVQA-manual

\section{Dataset Examples}
\label{sec:dataset_examples}

\cref{tab:suppl_dataset_fewshot_examples} shows 3 documents from HuDocVQA-manual and their corresponding human-annotated questions and answers. These 3 examples were used as few-shot exemplars to Llama 3.3 70B Instruct for generating HuDocVQA.

\begin{table*}[t]
\centering
\renewcommand{\arraystretch}{1.5}
\caption{Examples from our HuDocVQA-manual}
\label{tab:suppl_dataset_fewshot_examples}
\setlength{\tabcolsep}{8pt}
\begin{tabular}{|c|>{\raggedright\arraybackslash}p{5cm}|>{\raggedright\arraybackslash}p{5cm}|}
\hline
\textbf{Document} & \textbf{Question} & \textbf{Answer} \\
\hline
\hline
\raisebox{-\height}{\includegraphics[width=3cm]{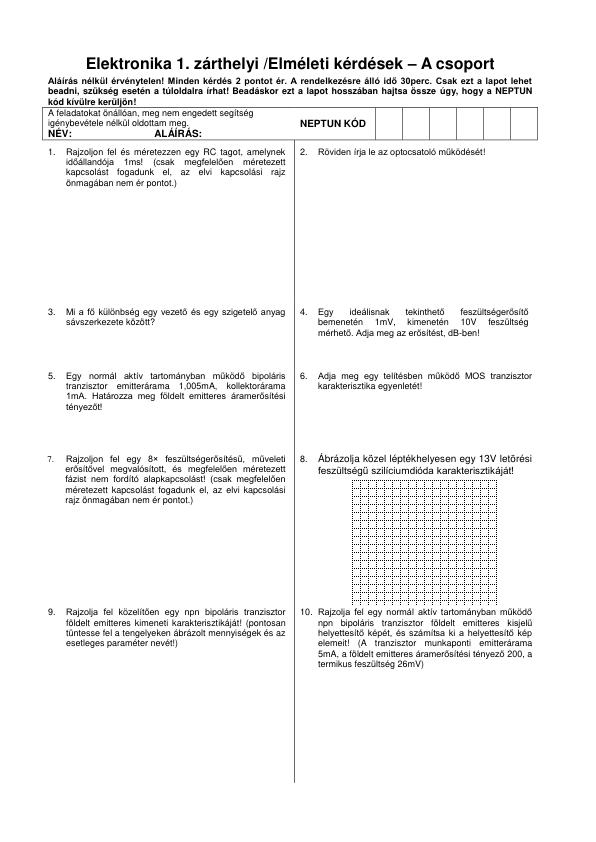}} & 
Mekkora a maximális feszültség, amelyen egy ideális feszültségerősítő mérhető? \textit{What is the maximum voltage at which an ideal voltage amplifier can be measured?} & 
Egy ideálisnak tekinthető feszültségerősítő bemenetén 1mV, kimenetén 10V feszültség , mérhető. \textit{An ideal voltage amplifier can measure 1mV at its input and 10V at its output.} \\
\hline
\hline 
\raisebox{-\height}{\includegraphics[width=3cm]{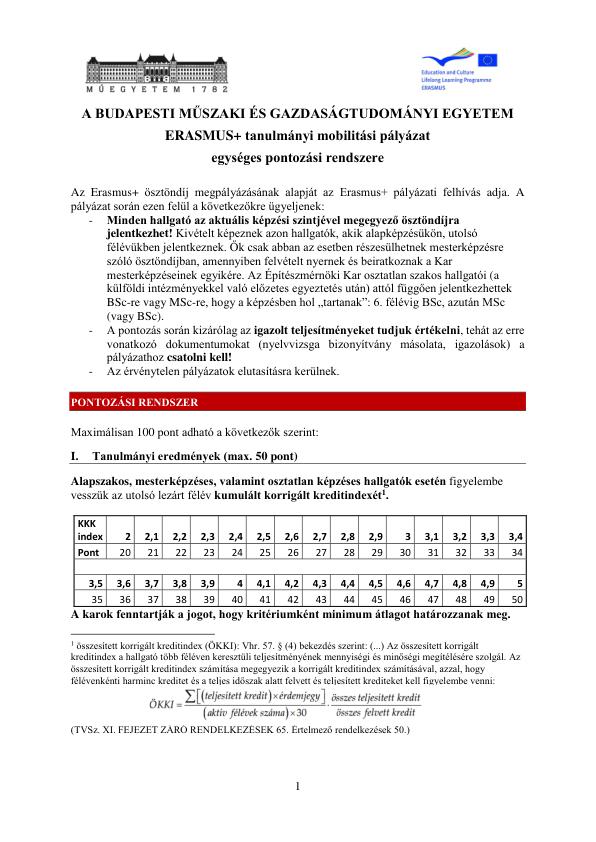}} & 
A legtöbb hallgató csak az aktuális iskolai végzettségének megfelelő ERASMUS+ ösztöndíjra pályázhat. Mi a kivétel? \textit{Most students can only apply for the ERASMUS+ scholarship corresponding to their current education level. What is the exception?} & 
Diákok, akik az utolsó félévben jelentkeznek alapképzésükre. \textit{Students who apply for their bachelor's degree in their last semester.} \\
\hline
\hline
\raisebox{-\height}{\includegraphics[width=3cm]{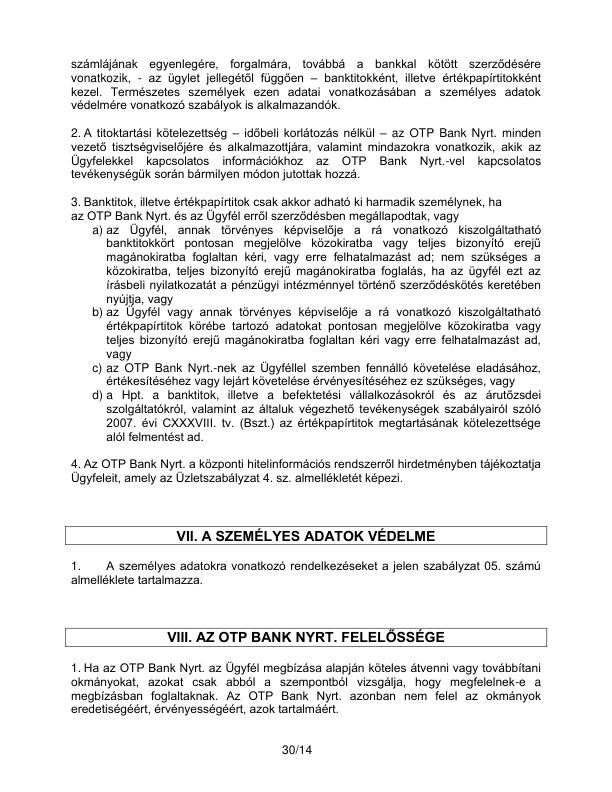}} & 
A személyes adatokra vonatkozó rendelkezéseket melyik almelléklet tartalmazza? \textit{Which sub-appendix contains the provisions on personal data?} & 
A személyes adatokra vonatkozó rendelkezéseket a jelen szabályzat 05. számú almelléklete tartalmazza. \textit{Provisions regarding personal data are contained in sub-appendix no. 05 of these regulations.} \\
\hline
\end{tabular}
\end{table*}

\section{Synthetic QA Generation Details and Prompt Format}
\label{sec:synqa_gen_details}

\begin{figure*}
    \centering
    \includegraphics[width=\textwidth]{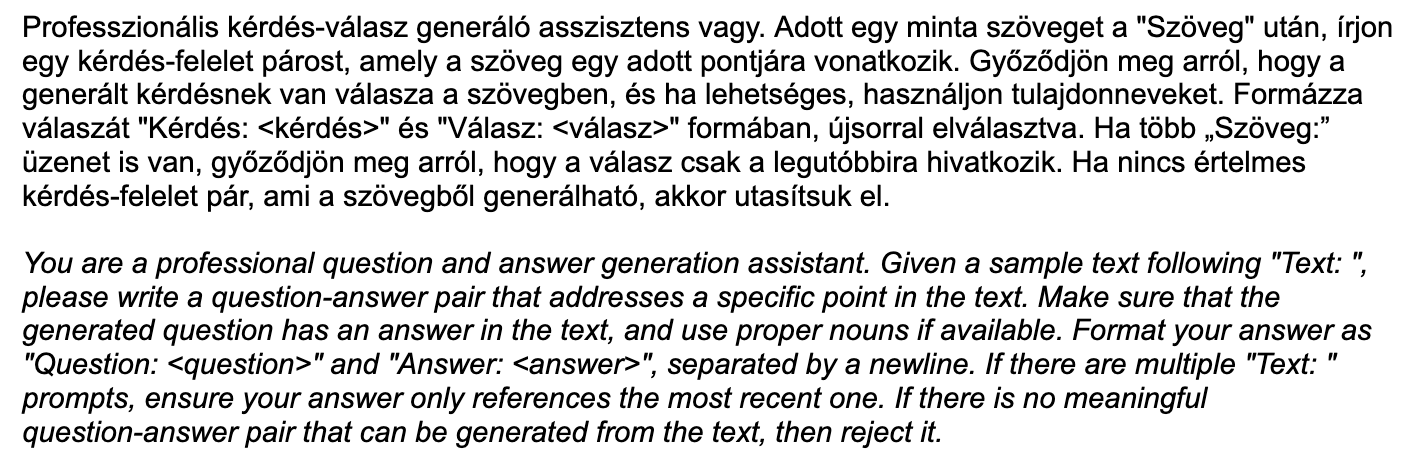}
    \caption{The Hungarian system prompt we provide to Llama 3.3 70B Instruct for synthetic QA generation. English translation provided in italics.}
    \label{fig:synqa_system_prompt}
\end{figure*}

When generating questions and answers with Llama 3.3 70B Instruct, we apply the system prompt in \cref{fig:synqa_system_prompt}. Then, after sampling a human-written input as a (text, question, answer) triple, we format a one-shot prompt as follows:

\begin{verbatim}
Szöveg: <text>
Kérdés: <question>
Válasz: <answer>
\end{verbatim}

Following this example, we then prompt the LLM with the text of the actual document we are annotating and the ``Kérdés: '' prompt. We sample a generation from the model with temperature 0.7, leaving all other sampling parameters default. If we can parse a newly generated question and answer from the model generation by splitting on the ``Kérdés: '' and ``Válasz: '' strings, we consider the generation successful and return the new question and answer. Otherwise, we discard the generation.

\section{Finetuning Details \& Hyperparameters}
\label{sec:train_hyperparam}

For all finetuning experiments, we train for 1 epoch and merge the original checkpoint with the last 3 checkpoints. \cref{tab:suppl_finetuning_hyperparams} shows other hyperparameters such as learning rate, batch size, etc. For all training experiments, we utilize LlamaFactory \cite{zheng2024llamafactoryunifiedefficientfinetuning} with 4xA100 80GB GPUs and PyTorch FSDP. We fully train all model parameters without using LoRA \cite{hu2021loralowrankadaptationlarge}.

\begin{table}[ht]
\centering
\renewcommand{\arraystretch}{1.4}
\begin{tabularx}{\linewidth}{|>{\raggedright\arraybackslash}X|>{\raggedright\arraybackslash}X|}
\hline
Peak Learning Rate & 1e-6 \\
\hline
LR Schedule & Cosine Decay \\
\hline
LR Warmup Ratio & 0.1 \\
\hline
Batch Size & 32 \\
\hline
Epochs & 1 \\
\hline
\end{tabularx}
\caption{Finetuning Hyperparameters}
\label{tab:suppl_finetuning_hyperparams}
\end{table}

\section{SFT Mixture}
\label{sec:suppl_sft_mixture}

\cref{tab:suppl_sft_mixture} outlines all datasets comprising our SFT mixture in \cref{tab:llama_finetuning}. As The Cauldron \cite{laurençon2024mattersbuildingvisionlanguagemodels}, DocMatix \cite{laurençon2024buildingbetterunderstandingvisionlanguage}, and LAION-COCO-NLLB \cite{visheratin2023nllbcliptrainperformant} each comprise several hundreds of thousands of examples, we subsample each dataset to 100k. For LAION-COCO-NLLB, we only subsample captions in Hungarian. As the HuggingFace implementation of Llama 3.2 does not natively handle multi-image inputs, we also filtered out multi-image examples from the Cauldron and DocMatix. The counts in \cref{tab:suppl_sft_mixture} reflect the number of single-image examples only.

\begin{table}[ht]
\centering
\renewcommand{\arraystretch}{1.4}
\begin{tabularx}{\linewidth}{|>{\raggedright\arraybackslash}X|>{\raggedright\arraybackslash}X|>{\raggedright\arraybackslash}X|}
\hline
\textbf{Dataset} & \textbf{\# Single-Image Examples} & \textbf{\# Subsampled}\\
\hline
The Cauldron \cite{laurençon2024mattersbuildingvisionlanguagemodels} & 1,643,352  & 100,000\\
\hline
DocMatix \cite{laurençon2024buildingbetterunderstandingvisionlanguage} & 565,009 & 100,000 \\
\hline
LAION-COCO-NLLB \cite{visheratin2023nllbcliptrainperformant} & 735,079 & 100,000 \\
\hline
HuDocVQA & 20,059 & 20,059 \\
\hline
\end{tabularx}
\caption{Datasets comprising our SFT mixture}
\label{tab:suppl_sft_mixture}
\end{table}

\section{Ablations}{
\label{sec:suppl_ablations}
\subsection{HuDocVQA Filtering Ablation}
\label{subsec:suppl_ft_ablations}

In \cref{tab:filtering_ablation} we finetune on HuDocVQA with and without the quality filters in \cref{subsubsec:quality_filters}. Although training on both datasets lead to degradation, adding the quality filters leads to an improvement of 1.9\%.

\begin{table}[ht]
\begin{tabularx}{\linewidth}{|>{\raggedright\arraybackslash}l|>{\centering\arraybackslash}X|}
    \hline
    \textbf{Model} & \textbf{HuDocVQA-manual}\\
    \hline
    Baseline & 0.481\\
    + unfiltered & 0.407 \\
    + filtered & 0.426\\
    \hline
  \end{tabularx}
  \caption{Ablation of quality filtering on HuDocVQA training.}
  \label{tab:filtering_ablation}
\end{table}

\section{Model Merging Ablation}
\label{sec:merge_ablation}

\cref{tab:merge_ablation} compares the accuracy of the final checkpoint versus a merged checkpoint for each training experiment on \cref{subsec:ft_experiments}. We apply simple model merging from \citet{wortsman2022modelsoupsaveragingweights} to Llama 3.1 11B Instruct and the last 3 checkpoints of each training run, where checkpoints are saved every 2000 steps.

\begin{table*}
  \centering
  \begin{tabularx}{\textwidth}{|l|X|X|}
    \hline
    \textbf{Finetuning Dataset} & \textbf{Final Checkpoint Accuracy}  & \textbf{Merged Checkpoint Accuracy}\\
    \hline
    + HuDocVQA & 0.269 & 0.285 \\
    + SFT mixture & 0.275 & 0.303 \\
    + SFT + 21k OCR & 0.333 & 0.374 \\
    + SFT + 105k OCR & 0.318 &  \textbf{0.404} \\
    \hline
  \end{tabularx}
  \caption{Ablations on using the final checkpoint versus merging the final 3 checkpoints. Accuracy is measured on the HuDocVQA test set using LLM-as-a-Judge}
  \label{tab:merge_ablation}
\end{table*}

\section{HuCCPDF Collection System Details}
\label{sec:pdf_collection}
Of the 3750 WARC files in CC 2021-17 we processed, we processed 25 at a time on CPU nodes with 200GB memory, with each WARC taking approximately 1.5 hours. In total, we downloaded 22 million PDFs, and filtered out 38,000 Hungarian PDFs according to the following criteria:


\begin{itemize}
  \item Checking if the PDF text from PyMUPDF was in Hungarian using \texttt{fasttext}\cite{fasttext}. We dropped any PDFs with a Hungarian language probability lower than 0.85. 
  \item Extracted text having fewer than 100 characters. Examples of this condition are pages that contain only an image, a blank page. Note this is a different threshold than the 60-character threshold referenced in \cref{tab:filters} for HuDocVQA
  \item Large mismatches in text and Markdown extracted from the PDF. PyMuPDF has functionalities for extracting both plaintext and Markdown. Such mismatches often indicated PDFs of low resolution or PDFs with complex tables that resulted in plaintext with low interpretability. 
  
  Specifically, for each page we calculated the absolute difference in whitespace-delimited word counts between plaintext and markdown. We discarded any pages where this difference exceeded 50\% of the plaintext word count.
\end{itemize}

\end{document}